%% file: root.tex
\documentclass{svproc}
\usepackage{url}

\usepackage{cite}
\usepackage{amsmath,amssymb,amsfonts}
\usepackage{algorithmic}
\usepackage{graphicx}
\usepackage{textcomp}
\usepackage{xcolor}
\usepackage{url}
\def\BibTeX{{\rm B\kern-.05em{\sc i\kern-.025em b}\kern-.08em
    T\kern-.1667em\lower.7ex\hbox{E}\kern-.125emX}}
    
\usepackage[linesnumbered,ruled]{algorithm2e}
\usepackage{xcolor}
\usepackage{paralist}
\newcommand{\norm}[1]{\left\lVert#1\right\rVert}

\let\labelindent\relax

\usepackage{amsthm}
\theoremstyle{plain}

\usepackage[linesnumbered,ruled]{algorithm2e}

\usepackage{dirtytalk}
\usepackage[caption=false]{subfig}
\usepackage{mathtools}

\begin{document}
\mainmatter              % start of a contribution
\title{Fast $k$-connectivity Restoration in Multi-Robot Systems for Robust Communication Maintenance}
\titlerunning{Fast $k$-connectivity Restoration}  % abbreviated title (for running head)
\author{Md Ishat-E-Rabban\inst{1} \and Guangyao Shi\inst{2} \and Griffin Bonner\inst{1} \and Pratap Tokekar\inst{1} }
\authorrunning{Rabban et al.} % abbreviated author list (for running head)
%\tocauthor{Ivar Ekeland, Roger Temam, Jeffrey Dean, David Grove, Craig Chambers, Kim B. Bruce, and Elisa Bertino}
%
\institute{University of Maryland, College Park MD 20742, USA\\
\email{ier,gbonner1,tokekar@umd.edu}\\
%\texttt{http://users/\homedir iekeland/web/welcome.html}
\and
University of Southern California, Los Angeles, CA 90007, USA\\
\email{shig@usc.edu}}

\maketitle              % typeset the title of the contribution

\begin{abstract}
Maintaining a robust communication network plays an important role in the success of a multi-robot team jointly performing an optimization task. A key characteristic of a robust cooperative multi-robot system is the ability to repair the communication topology in the case of robot failure. In this paper, we focus on the \textit{Fast $k$-connectivity Restoration} (FCR) problem, which aims to repair a network to make it $k$-connected with minimum robot movement. Here, a $k$-connected network refers to a communication topology that cannot be disconnected by removing $k-1$ nodes. We develop a Quadratically Constrained Program (QCP) formulation of the FCR problem, which provides a way to optimally solve the problem, but cannot handle large instances due to high computational overhead. We therefore present a scalable algorithm, called EA-SCR, for the FCR problem using graph theoretic concepts. By conducting empirical studies, we demonstrate that the EA-SCR algorithm performs within 10\% of the optimal while being orders of magnitude faster. We also show that EA-SCR outperforms existing solutions by 30\% in terms of the FCR distance metric.

\end{abstract}

\input{01_introduction}

\input{02_ralated_works}

\input{03_problem_formulation}

\input{04_qcp_formulation}
\input{05_heuristic_algorithm}

\input{06_experiments}

\input{07_conclusion}


\end{document}

%% file: 01_introduction.tex
\section{Introduction}
\label{intro}

There is a growing trend in assigning a team of robots to perform coverage, exploration, patrolling, and similar cooperative optimization tasks~\cite{rizk2019cooperative,yan2013a,robin2016multi,huang2019a,shi2021communication}. Compared to a single robot, a multi-robot system can extend the range of executable tasks, enhance task performance, and naturally provide redundancy for the system. Communication plays a key role in the successful deployment of a cooperative multi-robot system. To improve task performance, it is important to maintain uninterrupted communication among the robots. To ensure continuous information sharing and interaction, the robots in a team must always stay close to one another and maintain a connected communication topology among themselves.

However, maintaining a connected communication network may not always prove adequate for multi-robot systems, because robots may fail, e.g., experience sensor or actuator malfunction, or undergo an adversarial attack~\cite{zhou2021multi}. Any such issue may cause disconnections in the network, and consequently hurt task performance. Most existing works~\cite{engin2018minimiming,basu2004movement,abbasi2008movement,engin2021establishing,bredin2010deploying,atay2009mobile,luo2020minimally,luo2019minimum} on preventing such communication failures focus on maintaining a $k$-connected network among the robots. A \textit{k-vertex-connected} network is a topology that cannot be disconnected by removal of any combination of $k-1$ robots. Here, a high value of $k$ translates to enhanced robustness of the communication network, and vice versa.

\begin{figure}[!ht]
\centering
\includegraphics[width=1.0\linewidth]{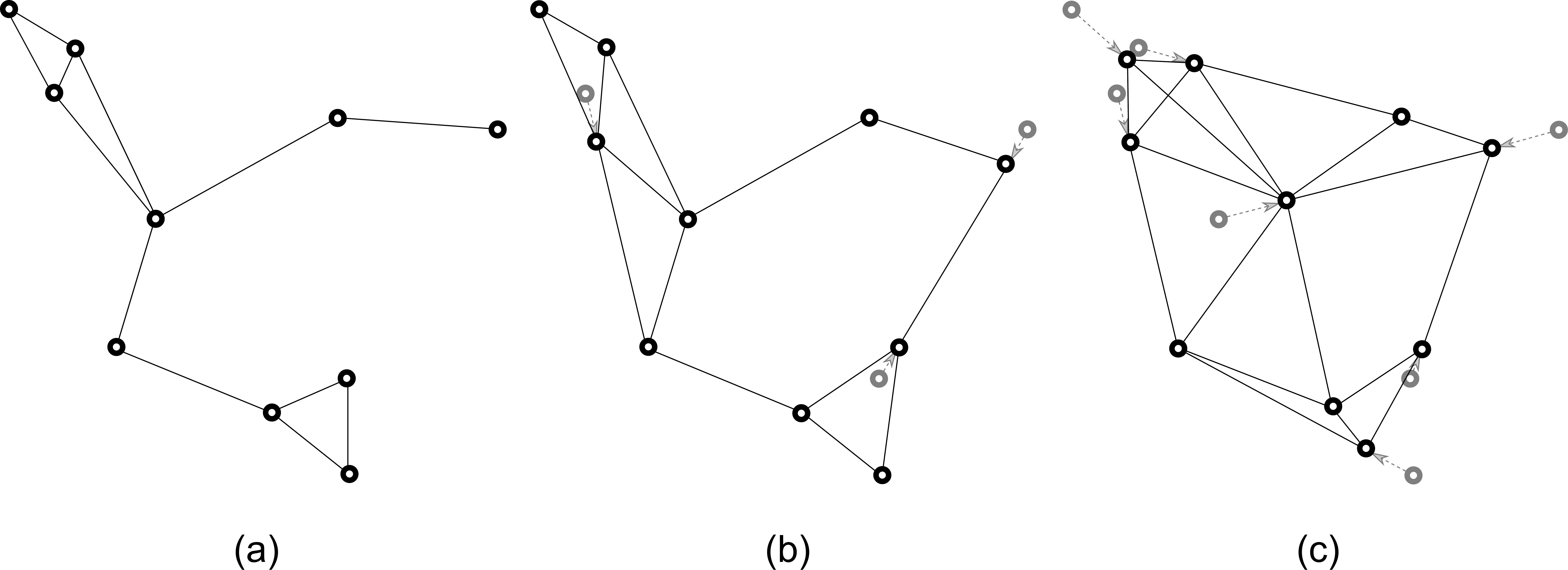}
\caption{(a) Initial 1-connected configuration. (b) Robot positions for $k=2$. (c) Robot positions for $k=3$. Dashed gray arrows show robot movements from initial positions.}
\label{figintro}
\end{figure}

In this paper, we investigate the problem of making a network $k$-connected while minimizing robot movements. The input to this problem is a set of robot positions and the value of $k$. Here, we assume a disk communication model with communication radius $h$. The goal is to find new positions of the robots which form a $k$-connected topology while minimizing the maximum distance between the previous and new positions of the robots, i.e., the \textit{minmax distance}. We call this problem the \textit{Fast k-connectivity Restoration} (FCR) problem. An example of the FCR problem instance is shown in Fig.~\ref{figintro}. The FCR problem and its variants have been extensively studied in literature~\cite{engin2018minimiming,basu2004movement,abbasi2008movement,engin2021establishing,bredin2010deploying}. Some of the above-mentioned variants of the FCR problem minimize different metrics, such as sum of robot movements, number of robots that need to be moved, etc., and some variants are only applicable for specific values of $k$. 

We develop a \textit{Quadratically Constrained Program} (QCP) formulation of the FCR problem, which is based on the idea of multi-commodity network flow~\cite{ilpref}. The QCP formulation enables us to solve the FCR problem optimally using a QCP solver. However, the QCP formulation can be used to solve only small instances of the FCR problem due to high computational overhead. 

%We call the QCP formulation-based optimal algorithm the OPT algorithm. 

We therefore propose a scalable algorithm, EA-SCR, to solve the FCR problem which requires less computational time than the optimal QCP-based algorithm. The EA-SCR algorithm solves the FCR problem by dividing it into two phases. In the first phase, we find a set of edges that, if augmented to the input network, make the network $k$-connected, and the maximum cost of the augmented edges is minimized. Here the cost of an edge is proportional to the distance between the two corresponding robots. We call this \textit{Graph Topology Optimization} (GTO) problem. We solve the GTO problem by first sorting the non-existent edges of the communication graph in increasing order of weight, and then keep adding the edges in order until the graph becomes $k$-connected. In the second phase, we move the robots in such a way that the edges obtained from the first phase is established, and the maximum distance traveled by a robot is minimized. We call this \textit{Movement Minimization} (MM) problem. To solve the MM problem, we propose a Breadth First Search (BFS) based idea called \textit{cascaded relocation}, which establishes the edges obtained from the first phase without removing any existing edges.

We conduct extensive experimentation to evaluate the performance of our proposed algorithm. The empirical results show that the EA-SCR algorithm generates solutions which are within 10\% of the optimal solution and gives 30\% lower minmax distance than all existing algorithms~\cite{basu2004movement,engin2021establishing}. We also test the EA-SCR algorithm in action by performing a hardware experiment using 6 drones.

In summary, we make the following contributions:
\begin{itemize}
    \item 
    We propose a scalable heuristic algorithm, EA-SCR, for the FCR problem. 
    \item
    We develop the first QCP formulation of the FCR problem which can be used to solve the FCR problem optimally.
    \item
    We conduct extensive experiments which demonstrate that the EA-SCR algorithm performs within 10\% of the optimal solution and outperforms all existing solutions by 30\% in terms of the minmax distance metric.
\end{itemize}

%% file: 02_ralated_works.tex
\section{Related Work}
\label{rel_work}

In a broader sense, this work is related to the connectivity problem in networked multi-robot systems. A large portion of research in this field uses algebraic graph theory to maintain the connectivity of the robotic network \cite{sabattini2013decentralized,robuffo2013passivity,zavlanos2011graph,sabattini2013distributed,minelli2020self, panerati2019robust}. These works are based on the fact that the second smallest eigenvalue of the Laplacian matrix of a graph, $\lambda_2$, is a measure of how well-connected a graph is. $\lambda_2$ is called \textit{algebraic connectivity} of a graph, which is 0 if the graph is not connected, and positive otherwise, with higher values representing better connectivity. Thus a gradient descent based controller can be used to move the robots in a direction that increases $\lambda_2$, and thus enhance the connectivity of a graph. However, algebraic connectivity is not the same as vertex connectivity~\cite{kirkland2002on}, and hence these methods cannot be used to guarantee $k$-vertex-connectivity. 

A number of works~\cite{demaine2009minimizing,anari2016euclinean,engin2018minimiming} study how to minimize movements to form 1-connected networks from arbitrary deployments. Demaine et al.~\cite{demaine2009minimizing} considered the case in which the robots are placed on the vertices of a graph and they can move only along the edges. Anari et al.~\cite{anari2016euclinean} proposed an algorithm with approximation ratio linear in the number of robots, $n$. Engin et al.~\cite{engin2018minimiming} considered the case in which the initial positions of the robots are chosen uniformly at random and they presented an algorithm with O($\sqrt{n}$)-approximation ratio.

Another body of work~\cite{basu2004movement,abbasi2008movement,lee2015connectivity,ramanathan2000topology} studies different variants of the 2-connectivity restoration problem. The work most closely related to the FCR problem was conducted by Basu et al.~\cite{basu2004movement}. In this work, the authors proposed a Block Cut Tree based algorithm to make a 1-connected network 2-connected and their objective was to minimize the sum of the movements of all the robots. Abbasi et al.~\cite{abbasi2008movement} considered the case in which 2-connectivity is lost due to the removal of one robot and proposed an algorithm that moves minimum number of robots to restore 2-connectivity. Lee et al.~\cite{lee2015connectivity} considered the same objective as us, but they assumed that additional robots can be used to facilitate the transition to a 2-connected topology. Ramanathan et al.~\cite{ramanathan2000topology} addressed a variant in which the goal is to achieve a 2-connected network while minimizing the maximum level of power assignment. These works are only applicable for 2-connecting a network.

$k$-connectivity maintenance and restoration in multi-robot systems has been extensively studied in literature~\cite{engin2021establishing,bredin2010deploying,atay2009mobile,luo2020minimally,luo2019minimum}. This paper is most closely related to the work of Engin et al.~\cite{engin2021establishing}, who proposed an algorithm for the FCR problem which first forms a $k+1$-clique at the center of the data-space, and incrementally adds the rest of the robots without compromising $k$-connectivity. In~\cite{bredin2010deploying,atay2009mobile}, an FCR variant was considered which allows introducing new robots in the network, and the goal was to achieve $k$-connectivity by deploying the smallest number of additional robots. Luo et al.~\cite{luo2019minimum,luo2020minimally} considered the problem of minimizing the sum of deviations from a prescribed controller while maintaining $k$-connectivity. 

Distributed algorithms~\cite{cornejo2010fault,akram2020distributed,abbasi2008movement} to maintain and restore $k$-connectivity have been well-studied by researchers. Cornejo et al.~\cite{cornejo2010fault} and Akram et al.~\cite{akram2020distributed} propose distributed algorithms for $k$-connectivity maintenance. In these methods, the robots rely on local information from up to 2-hop neighbors to test for $k$-connectivity. Consequently these algorithms require less computation than global and centralized solutions. But, these methods are not guaranteed to return a correct solution and may result in false positives. Abbasi et al.~\cite{abbasi2008movement} proposed a distributed algorithm to restore 2-connectivity, but their algorithm can only handle the cases in which 2-connectivity is lost due to failure of a single node.

%% file: 03_problem_formulation.tex
\section{Problem Formulation }
\label{pform}

First, we introduce some graph-theoretic concepts. A graph $G$ is a tuple $(V_G, E_G)$, where $V_G$ and $E_G$ are the set of vertices and edges of $G$ respectively. A graph $G$ is \textit{connected}, if there exists a path between each pair of vertices in $G$. A graph $G$ is \textit{k-connected} (or \textit{k-vertex-connected}), if for each $v \subseteq V_G$ with $|v|=k-1$, the graph $G-v$ is connected. Here, $|.|$ denotes set cardinality and $G-v$ is the graph obtained by removing from $G$ the vertex $v$ and all edges incident on $v$. 

We assume that $n$ robots are operating in an unobstructed 2D or 3D environment. We denote the $i^{th}$ robot by $r_i$, and the position of $r_i$ by $x_i$, where $x_i$ is a 2D or 3D point. $R$ represents the set of robots, $R=\{r_1,r_2,\ldots, r_n\}$, and $X$ represents the set of positions of the robots, $X=\{x_1,x_2,\ldots, x_n\}$. 

Given the robot positions $X$ and the communication radius $h$, we define the \textit{communication graph} $G(X)$ as follows. $G(X)$ is an unweighted graph which has one vertex corresponding to each robot and there exists an edge between a pair of vertices if the corresponding two robots are within the communication range of each other, i.e., if the Euclidean distance of the two robots is at most $h$. 

\textbf{Problem 1 (Fast k-connectivity Restoration):} We are given a set of positions $X$ of $n$ robots, the communication radius $h$, and the degree of connectivity $k$. The FCR problem aims to find new positions $X^*=\{x_1^*,x_2^*,\ldots ,x_n^*\}$ of the robots such that $G(X^*)$ is $k$-connected and the maximum movement of the robots is minimized. Mathematically, 
\begin{equation}
    \min \max_{1 \leq i \leq n} \norm{x_i^*-x_i} \quad
     s.t. \, \, \,  G(X^*)\, \, {\rm is \, \, \text{$k$-connected}}
\end{equation}

Here, $\norm{.}$ denotes euclidean distance. We assume that the robots have single-integrator dynamics and operate at constant speeds. Therefore, the time a robot takes to transit between two points is proportional to the corresponding distance. It should be noted that we abstract away collision avoidance from the formulation due to the fact that the communication radius of robots is usually much larger than the size of robots \cite{kuzminykh2017testing, iordache2017field}. In such cases, the influence of collision avoidance on the controllers can be ignored.

Now we outline the organization of the rest of the paper. In Section~\ref{qcpform}, we present the QCP formulation of the FCR problem. In Section~\ref{heuralgo}, we describe the EA-SCR algorithm. In Section~\ref{exp}, we present the experimental findings. 

%We assume that $n$ robots are operating in an unobstructed 2D or 3D environment. Let $\mathcal{R}=\{r_1, \ldots, r_n\}$ be the set of all robots. We denote the position of the robot $r_i$ by $x_i$, where $x_i$ is a 2D or 3D point. $X$ denotes the set of positions of the robots, $X=\{x_1,x_2,\ldots, x_n\}$. Given robot positions $X$ and communication radius $h$, we construct a \textit{communication graph} as follows. A robot corresponds to a vertex $i \in {V}$ and $(i,j) \in {E}$ if $\norm{x_i-x_j} \leq h$, where $\norm{.}$ denotes euclidean distance. We use $G(X)$ to denote the communication graph induced by $X$. Robots have single-integrator dynamics and operate  at  maximum  speeds. Therefore, the time that it takes to transit between two points is proportional to the corresponding distance. In the rest of this paper, we will use the maximum moving distance to denote the maximum transition time. {It should be noted that we abstract away collision avoidance from the formulation due to the fact that the communication radius of robots is usually much larger than the size of robots \cite{kuzminykh2017testing, iordache2017field} and they need to augment the network to be 2-connected only when they are already quite far from each other. In such cases, the influence of behavior controllers for collision avoidance can be ignored.}

%% file: 04_qcp_formulation.tex
\section{QCP Formulation }
\label{qcpform}

In this section, we present a Quadratically Constrained Program (QCP) formulation, which can be used to find the optimal solution to the FCR problem. The key idea behind the QCP formulation is based on multi-commodity network flow~\cite{ilpref}. In the QCP formulation, we use the fact that if a graph is $k$-connected, there exists at least $k$ vertex-disjoint paths between each pair of vertices. We use the following variables in the QCP formulation.

\begin{itemize}
    \item Binary variable $e_{i,j}$, where $1\leq i,j\leq n$, indicates if there exists an edge between robots $x_i^*$ and $x_j^*$ in the communication graph of $X^*$ with respect to the communication radius $h$, i.e., if $x_i^*$ and $x_j^*$ are within distance $h$ of each other. 
    
    \item Binary variable $z_{s,d,i,j}$, where $1\leq s,d,i,j\leq n, \, \, s\neq d$, indicates if the edge between robots $r_i$ and $r_j$ is included in a vertex-disjoint path from source robot $r_s$ to destination robot $r_d$. 
    
    \item $x_i^*$, where $1\leq i\leq n$, represents a tuple of real-valued variables (x, y, and optionally z coordinates), which indicates the new position of the $i^{th}$ robot.
    
    \item Real valued variable $z^*$ is used to find the maximum movement of the robots.
    
\end{itemize}

The QCP formulation is presented below. We use the objective function and the set of constraints in Equation (\ref{ilpc1}) to minimize the maximum movement of the robots. Constraints (\ref{ilpc2}) ensure that if a pair of robots are within the communication radius of each other, the corresponding $e$ variable is set to 1, and vice versa. Here, M is a large positive constant. Note that, we manually set $e_{i, i}=0$, for $1\leq i\leq n$, to indicate that there are no self edges. Constraints (\ref{ilpc3}) enforce that only valid edges (i.e., edges between robots within each other's communication radius) are used to form source-destination paths. Constraints (\ref{ilpc4}) are used to maintain flow conservation. In other words, for each source-destination pair, the outgoing flow should be $k$ greater than the incoming flow for the source vertex, the outgoing flow should be $k$ less than the incoming flow for the destination vertex, and incoming and outgoing flow should be equal for all the other vertices. Constraints (\ref{ilpc5}) ensure vertex-disjointedness, i.e., for each source-destination pair, an internal node is included in at most one path.

$$\min  z^* $$
\begin{equation}\tag{2}
\label{ilpc1}
\text{s.t.} \norm{x_i^*-x_i}^2 \leq z^* \quad \quad  \forall \,  \, 1 \leq i \leq n 
\end{equation}  

\vspace{-0.4cm}

\begin{equation}\tag{3}
\label{ilpc2}
-{\rm M} e_{i,j} \leq \norm{x_i^*-x_j^*}^2 - h^2 \leq {\rm M}(1-e_{i,j})  \quad  \forall \, \, 1 \leq i \neq j \leq n 
\end{equation}

\vspace{-0.3cm}

\begin{equation}\tag{4}
\label{ilpc3}
z_{s,d,i,j} \leq e_{i,j}  \quad \quad \forall  \, 1 \leq s,d,i,j \leq n 
\end{equation}

\vspace{0.1cm}

$\forall \, \, 1 \leq s,d,i \leq n, \, \, \, {\rm such \, \, that,} \, s \neq d$,
\vspace{-0.1cm}
\begin{equation}\tag{5}
\label{ilpc4}
    \sum_{1 \leq j \leq n} z_{s,d,i,j} - \sum_{1 \leq j \leq n} z_{s,d,j,i} = 
\begin{cases}
    k, \; {\rm if} \, \, i = s\\
    -k, \; {\rm if} \, \, i = d\\
    0, \; {\rm otherwise} 
\end{cases}
\end{equation}

$\forall \, \, 1 \leq s,d,i \leq n, \, \, \, {\rm such \, \, that,} \, i \neq s, i \neq d, s \neq d$,
\begin{equation}\tag{6}
\vspace{-0.1cm}
\label{ilpc5}
    \sum_{1 \leq j \leq n} z_{s,d,i,j} \leq 1
\end{equation}

Note that, the QCP formulation has O($n^4$) binary variables. Hence it can be used to solve only very small instances of the FCR problem. To this end, in the following section, we present an algorithm that finds approximate solutions to the FCR problem with less computational overhead.  

%% file: 05_heuristic_algorithm.tex
\section{Scalable Algorithm: EA-SCR}
\label{heuralgo}

In this section, we describe our main contribution, the EA-SCR algorithm, which provides a more scalable alternative compared to a QCP-based solution for the FCR problem. We divide the FCR problem into two sub-problems, namely, the Graph Topology Optimization (GTO) problem, and the Movement Minimization (MM) problem. The GTO problem is to determine which edges to augment to make a graph $k$-connected and the MM problem is to find out how to realize those edges such that maximum robot movement is minimized.

\begin{algorithm}[htp]
\label{eascralgo}
\caption{EA-SCR}
    \SetKwInOut{Input}{Input}
    \SetKwInOut{Output}{Output}
    \Input{Set of robot positions $X$, Communication radius $h$,\\ Degree of connectivity $k$}
    \Output{New set of robot positions $X^*$}
    \SetAlgoLined
    \DontPrintSemicolon
    \SetKwFunction{SCR}{SCR}
    \SetKwFunction{EA}{EA}
    \SetKwFunction{EASCR}{EASCR}
    
    \SetKwProg{myalg}{Algorithm}{}{}
    \myalg{\EASCR{$X, h, k$}}
    {
        $E_a \leftarrow$ \EA{$X,h,k$}\;
        Return \SCR{$X,h,E_a$} \;   
    }
\end{algorithm}  

Now we provide a high-level overview of the EA-SCR algorithm, which has two steps. First we solve the GTO problem using the Edge Augmentation (EA) algorithm (described in Section~\ref{gtosec}) to determine the set of edges $E_a$ to be augmented. Next we solve the MM problem using the Sequential Cascaded Relocation (SCR) algorithm (described in Section~\ref{mmsec}) to determine where to move the robots such that the edges obtained from the EA algorithm are established while the maximum robot movement is minimized. The pseudo-code of the EA-SCR algorithm is provided in Algorithm~\ref{eascralgo} above.

%Now we summarize the two steps of the EA-SCR method in Algorithm~\ref{eascralgo}. First we determine the augmentation set using the EA algorithm (Algorithm~\ref{algorithm:edge_augmentation}) and then we feed the augmentation set to the SCR algorithm (Algorithm~\ref{scralgo}) to determine the new positions of the robots.

\subsection{Graph Topology Optimization}
\label{gtosec}

In the GTO problem, we are given a set of robot positions $X$, the communication radius $h$, and the degree of connectivity $k$. The GTO problem aims to determine a set of edges $E_a$ which are not present in $G(x)$, i.e. $E_a \subseteq \overline{E_{G(X)}}$, such that the graph ($V_{G(X)}, E_a \cup E_{G(X)}$) is $k$-connected and the weight of the most costly edge in $E_a$ is minimized. Here, $\overline{A}$ denotes the complement of set $A$ and the weight of an edge between robots $r_i$ and $r_j$ in $\overline{E_{G(X)}}$ is $w(i,j)=\norm{x_i-x_j}$. In other words, the GTO problem aims to find a set of edges, $E_a$, such that augmenting the edges of $E_a$ makes $G(X)$ $k$-connected and the weight of the costliest edge in $E_a$ is minimized. We call $E_a$ the \textit{Augmentation Set}.

\subsubsection{Edge Augmentation Algorithm}
\label{eaalgo}

\begin{algorithm}[htp]
\label{algorithm:edge_augmentation}
\caption{Edge Augmentation}
    \SetKwInOut{Input}{Input}
    \SetKwInOut{Output}{Output}
    \Input{Set of robot positions $X$, Communication radius $h$, \\Degree of connectivity $k$}
    \Output{Augmentation set $E_a$}
    \SetAlgoLined
    \DontPrintSemicolon
    \SetKwFunction{EA}{EA}
    \SetKwProg{myalg}{Algorithm}{}{}
    \myalg{\EA{$X, h, k$}}
    {
        compute $G(X)$ \;
        $E \gets \overline{E_{G(X)}}$ \;
        sort($E$) \;
        $count \gets 0$ \;
        \While{$G(X)$ is not k-connected}
        {
            $G(X)$.addEdge($E[count]$)\;
            $count \leftarrow count + 1$\;
        } 
        $E_a \gets \emptyset $ \;
        \For{$i \leftarrow 0 \, \, to \, \, count-1 $}
        {
            $G(X)$.removeEdge($E[i]$)\;
            \If{$G(X)$ is not k-connected}
            {
                $G(X)$.addEdge($E[i]$)\;
                $E_a$.insert($E[i]$)\;         
            } 
        }
        Return $E_a$ \;
    }
\end{algorithm} 

In this section, we present the \textit{Edge Augmentation} (EA) algorithm (Algorithm~\ref{algorithm:edge_augmentation}) which determines the augmentation set. In the EA algorithm, first we sort the edges not present in $G(X)$ in increasing order of weight (Lines 2-4). Next we keep adding the edges in the sorted order one after another to $G(X)$  until $G(X)$ becomes $k$-connected (Lines 5-9). Finally, we remove those edges from $G(X)$ whose removal does not compromise the $k$-connectivity of $G(X)$ (Lines 11-17). The proof of correctness of Algorithm~\ref{algorithm:edge_augmentation} follows by construction, as the for loop in Lines 11-17 ensures that all edges contributing to $k$-connectivity is inserted into the returned edge set $E_a$. 

The running time of the EA algorithm depends on the value of $k$, because the time complexity of testing $k$-connectivity of a graph (in Line 6 and Line 13) depends on the value of $k$. For $k=1$ and $k=2$, $k$-connectivity can be tested in linear time~\cite{bicontest} by performing a Depth First Search (DFS). For higher values of $k$, we can test for $k$-connectivity using tests for $k-1$-connectivity. For example, in the case of $k=3$, we remove one vertex $v$ from the graph and test for 2-connectivity of the remaining graph using the DFS based method. If, for all $v$, the graph remains 2-connected after removal of $v$, the graph is 3-connected and vice versa. Also, generalized $k$-connectivity test can be performed using a maximum flow based algorithm~\cite{kcontest}.

%but this algorithm has a prohibitively high time complexity of O($n^5)$. 

\subsection{Movement Minimization}
\label{mmsec}

In the MM problem, we are given a set of robot positions $X$, the communication radius $h$, and the augmentation set $E_a$. The goal of the MM problem is to find new positions $X^*=\{x_1^*,x_2^*,\ldots ,x_n^*\}$ of the robots such that $E_a \cup E_{G(X)} \subseteq E_{G(X^*)}$ and the maximum movement of the robots is minimized. Mathematically,

\begin{align*} 
    \min \max_{1 \leq i \leq n}  \norm{x_i^*-x_i}  \; \; \; \;  \text{s.t.} \; \norm{x_i^*-x_j^*} \leq h, \; \; \; \forall (i,j) \in E_a \cup E_{G(X)}.
\end{align*}

In other words, the MM problem aims to move the robots in such a way that the existing edges of $G(X)$ are retained, and additionally, each pair of robots in $E_a$ comes within the communication radius of each other, while the maximum movement of the robots is minimized.

\subsubsection{Sequential Cascaded Relocation Algorithm}
\label{scrsec}

%, and $A \backslash B$ denotes the set of elements in $A$ that are not in $B$

In this section, we present the \textit{Sequential Cascaded Relocation} (SCR) algorithm (Algorithm~\ref{scralgo}) for the MM problem. The SCR algorithm is based on the idea of \textit{Cascaded Relocation} (CR). A CR refers to moving one vertex of a graph to a new location while retaining the existing edges of the graph. Note that, moving only one vertex may disconnect some of the existing edges of the graph. To ensure that the existing edges are retained, a series of relocations may become necessary, hence the process is named cascaded relocation. 

The CR process is described in the Procedure \textit{relocate} in Algorithm~\ref{scralgo}. Here \textit{relocate(X,h,i,j,d)} refers to the movement of the $i^{th}$ robot towards the direction of the $j^{th}$ robot by a distance of $d$ without disconnecting any existing edge of $G(X)$, where $X$ and $h$ have their usual meaning. To achieve this, first we perform a Breadth-First Search on $G(X)$ rooted at the $i^{th}$ robot, $r_i$ (Line 9). Performing BFS provides the graph distance of all the other robots from $r_i$ and the parent of all the other robots on the BFS-tree rooted at $r_i$. The parent of robot $k$ is denoted as $par(k)$. Next we move robot $i$ to its goal location (Line 10). Then we consider all the other robots in increasing order of graph distance from $r_i$ (Line 11). If a robot gets disconnects from its parent due to the relocation of the parent, we move the robot minimally towards the parent such that the disconnected edge is reestablished (Line 12-14).

\begin{algorithm}[htp]
\label{scralgo}
\caption{Sequential Cascaded Relocation}
    \SetKwInOut{Input}{Input}
    \SetKwInOut{Output}{Output}
    \Input{Set of robot positions $X$, Communication radius $h$,\\Augmentation set $E_a$}
    \Output{New set of robot positions $X^*$}
    \SetAlgoLined
    \DontPrintSemicolon
    \SetKwFunction{SCR}{SCR}
    \SetKwFunction{relocate}{relocate}
    \SetKwFunction{BFS}{BFS}
    
    \SetKwProg{myalg}{Algorithm}{}{}
    \myalg{\SCR{$X, h, E_a$}}
    {
        \For{$(i, j) \in E_a$}
        {
            $d \leftarrow \norm{x_i - x_j} - h$\;
            \relocate{$X, h, i, j, \frac{d}{2}$}\;
            %$d \leftarrow \norm{x_i - x_j} - h$\;
            \relocate{$X, h, j, i, \frac{d}{2}$}\;
        }  
        Return $X$ \;
    }

    \SetKwProg{myproc}{Procedure}{}{}
    \myproc{\relocate{X, h, i, j, d}}
    {
        $G(X).\BFS{i}$ \;
        $x_i =  x_i + d \frac{x_j - x_i}{\norm{x_j - x_i}}$ \;
        \For{$k \in \{1, \ldots n \} \backslash \{ i \}$ \normalfont{in increasing order of graph distance from} $i$}
        {
            $ d' \leftarrow \norm{x_k - x_{par(k)})} - h$\;
            \If{$d' > 0$}
            {
                $ x_k =  x_k + d' \frac{x_{par(k)} - x_k}{\norm{x_{par(k)} - x_k}}$
            } 
        }
    }
\end{algorithm}

\begin{figure}[!ht]
\centering
\includegraphics[width=0.6\linewidth]{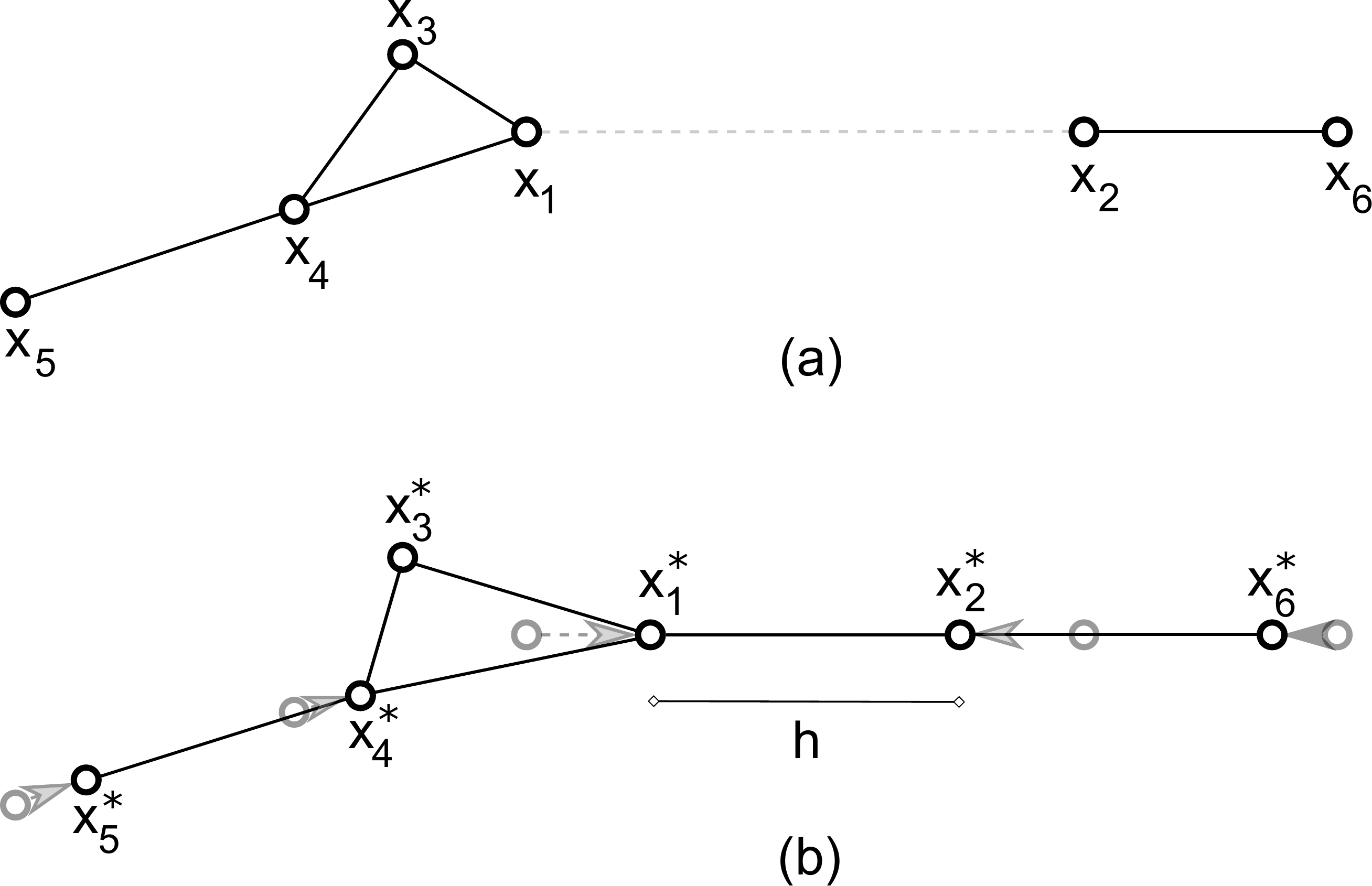}
\caption{Before (a) and after (b) connecting edge (1,2). First, $r_1$ moves towards $r_2$ by distance $\frac{d}{2}$, where $d=\norm{x_1 - x_2}-h$. This movement disconnects edge (1,4), hence $r_4$ moves minimally towards $r_1$ to reconnect the edge. Subsequently edge (4,5) gets disconnected and $r_5$ moves minimally towards $r_4$ to reconnect the edge. Finally, $r_2$ moves towards $r_1$ by distance $\frac{d}{2}$ connecting (1,2). This disconnects edge (2,6), hence $r_6$ moves towards $r_2$ to reconnect the edge. Gray arrows in (b) show robot movements.}
\label{figscr}
\end{figure}

Now we describe the SCR algorithm (Algorithm~\ref{scralgo}), where we augment the edges in the augmentation set to the communication graph by performing a sequence of CRs. In the SCR algorithm, we perform two CRs to connect each edge of the augmentation set. For each edge $(i,j)$ in $E_a$, first we compute the minimum movement $d$ required to connect $r_i$ and $r_j$ (Line 3). Then we move $r_i$ and $r_j$ towards each other by half the distance of $d$ using the \textit{relocate} method (Line 4-5), and thus establish the edge $(i,j)$. A demonstration of how one edge is established by the SCR method is shown in Figure~\ref{figscr}. The proof of correctness of the SCR algorithm follows by construction. The running time of the \textit{relocate} procedure is linear as BFS takes linear time. Thus the time complexity of the SCR algorithm is O($|E_a|(n+m)$), where $m$ is the number of edges in $G(X)$.

%Now we summarize the two steps of the EA-SCR method in Algorithm~\ref{eascralgo}. First we determine the augmentation set using the EA algorithm (Algorithm~\ref{algorithm:edge_augmentation}) and then we feed the augmentation set to the SCR algorithm (Algorithm~\ref{scralgo}) to determine the new positions of the robots.

%Please refer to~\cite{abbasi2008movement} for the proof that the \textit{relocate} routine does not disconnect any existing edges of $G(X)$. Thus the proof of correctness of the SCR algorithm follows by construction. We also propose a QCP formulation to optimally solve the MM problem, which is described in Appendix~\ref{optmmsec}. We call the FBR algorithm that uses the SCR algorithm and QCP formulation to solve the MM problem the EA-SCR and EA-OPT algorithm respectively.

%% file: 06_experiments.tex
\section{Experiments}
\label{exp}

In this section, first we describe the experimental setup (Section~\ref{expsetup}), next we present the empirical results (Section~\ref{empres}), and finally we discuss a hardware experiment deploying the EA-SCR algorithm (Section~\ref{pmcf}).  

\subsection{Experimental Setup}
\label{expsetup}

\textbf{Compared Algorithms}: We empirically compare the performance of two algorithms proposed in this paper (OPT and EA-SCR) and two existing algorithms (NB~\cite{engin2021establishing}, and BT~\cite{basu2004movement}). We use a commercial QCP solver, Gurobi~\cite{grb}, to optimally solve the QCP presented in Section~\ref{qcpform}, which we call the OPT algorithm. The EA-SCR algorithm is described in Section~\ref{heuralgo}.

Engin et al.~\cite{engin2021establishing} proposed the \textit{Net-Builder} (NB) algorithm to solve the FCR problem. In the NB algorithm, first $k+1$ robots located closest to the center of the data space are moved using a geometric transformation called homothety to form a $k+1$-clique. In the next phase, the remaining robots are added to the clique in increasing order of distance such that $k$-connectivity is preserved.

The \textit{Block Translation} (BT) algorithm proposed by Basu et al.~\cite{basu2004movement} is based on Block Cut Trees and is applicable for $k=2$. In each iteration of this algorithm, the leaf blocks of the Block Cut Tree are translated towards its parent node such that the leaf block merges with the parent block. This process is repeated until there is one block left, i.e., the graph becomes 2-connected. 

%Abbasi et al.~\cite{abbasi2008movement} presented the \textit{Distributed Actor Relocation} (DAR) algorithm. The DAR algorithm is applicable for $k=2$, and the input must be a connected graph that results from the removal of a vertex from a 2-connected graph. In this algorithm, the closest neighbor of the removed vertex is moved towards the position of the removed vertex. If the movement disconnects any existing edges, a cascaded relocation like procedure is used to restore those edges. 

\textbf{Dataset}: We use two types of 2D synthetic datasets: uniform and GMM (Gaussian Mixture Model~\cite{gmm}). First, we generate random 2D points representing the positions of the robots uniformly (or, according to a GMM) for uniform (or, GMM) dataset. Next, we construct the communication graph of the generated points using $h=1m$. If the communication graph is connected, we add the set of points to our dataset. We do not add disconnected topologies to our dataset, because the BT algorithm requires the graph to be connected. Our experiments show that the performance of the compared algorithms are similar for both datasets. Hence, for brevity, we show the results only for uniform dataset. 

%For experiments with the DAR algorithm, we use connected graph topologies that can be made 2-connected by adding one robot. We call this the DAR dataset.

\textbf{Evaluation Metric}: We use three metrics to empirically evaluate our proposed algorithms: minmax distance, total distance, and running time. The objective of the FCR problem is to minimize the maximum distance between the previous and new positions of the robots. We call this metric the minmax distance. The total distance is the sum of the movements of all the robots.

\textbf{Platform}: The algorithms are implemented using C++. The experiments are conducted on a core-i7 2.7GHz PC with 16GB RAM, running Microsoft Windows 11.

\subsection{Empirical Results}
\label{empres}

\begin{figure}[!ht]
\centering
\includegraphics[width=0.70\linewidth]{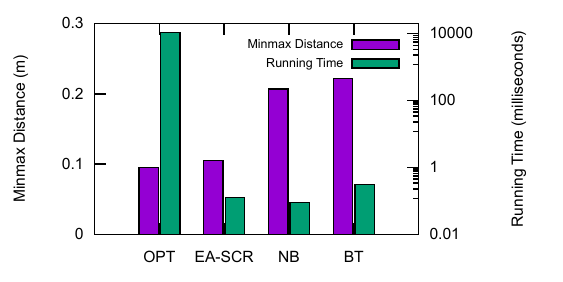}
\caption{Comparison with OPT algorithm using $n=8$ and $k=2$.}
\label{figc}
\end{figure}

\textbf{Comparison with OPT}: In this experiment, we compare the minmax distance and running time of the discussed algorithms. As the OPT algorithm can be used to solve only small instances of the FCR problem, we use 8 robots in this experiment and the value of $k$ is set to 2. The experimental results in Figure~\ref{figc} show that the OPT algorithm requires 5 orders of magnitude higher running time than the other algorithms (because of the high complexity of the QCP formulation). In terms of minmax distance, our proposed EA-SCR algorithm performs within 10\% of the OPT algorithm and almost 50\% better than the existing algorithms (NB and BT). We do not report results for the total distance metric because our QCP formulation optimizes the minmax distance; not the total distance. In all the experiments, we use a communication radius of $1m$. We perform each experiment 100 times and report the average. 

\begin{figure}[!ht]
\centering
\includegraphics[width=1.0\linewidth]{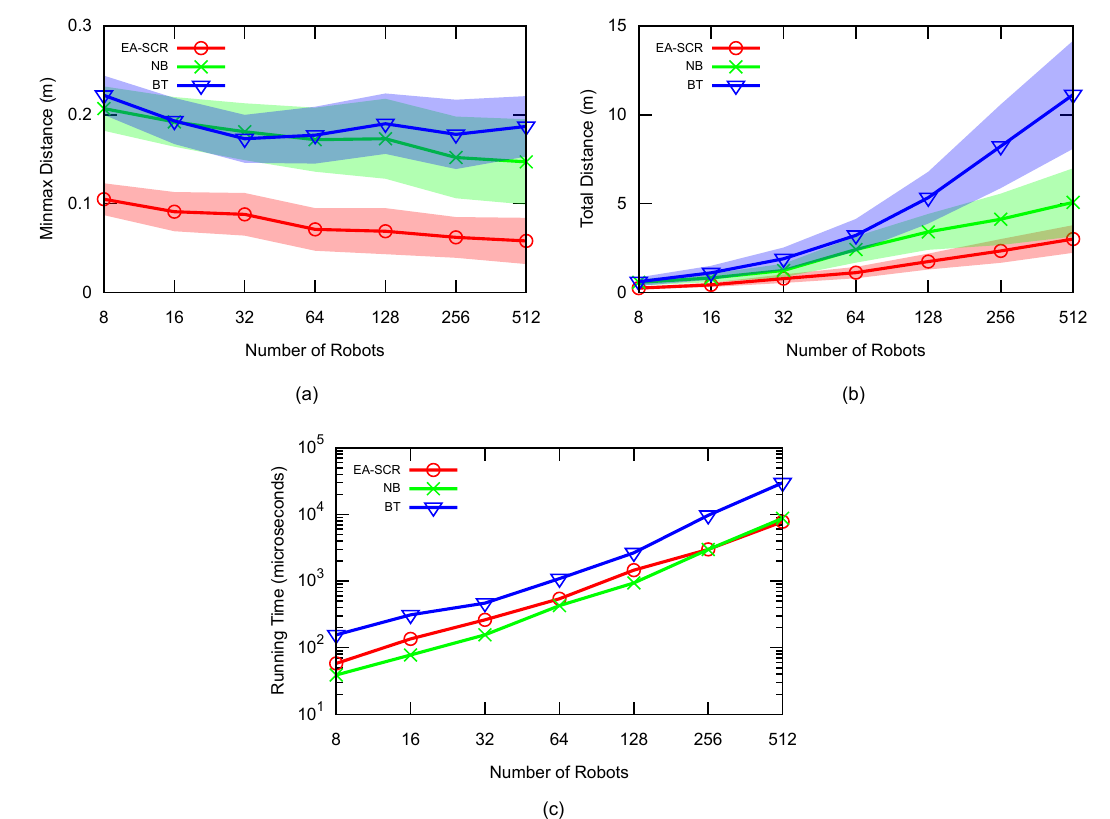}
\caption{Comparison of EA-SCR, NB, and BT algorithms for $k=2$ in terms of (a) minmax distance, (b) total distance, and (c) running time.}
\label{figd}
\end{figure}

\textbf{Results for k=2}: In this experiment, we compare the discussed algorithms (except for OPT) in terms of minmax distance, total distance, and running time for $k=2$. We vary the number of robots from 8 to 512 in increments by a factor of 2. The experimental results in Figure~\ref{figd} show that our proposed EA-SCR algorithm generates solutions with 50\% lower minmax distance and 30\% lower total distance compared to the NB and BT algorithms. The performance of NB and BT are equivalent according to minmax distance, and NB performs better than BT in terms of total distance. Here, the shades show standard deviation.

\begin{figure}[!ht]
\centering
\includegraphics[width=1.00\linewidth]{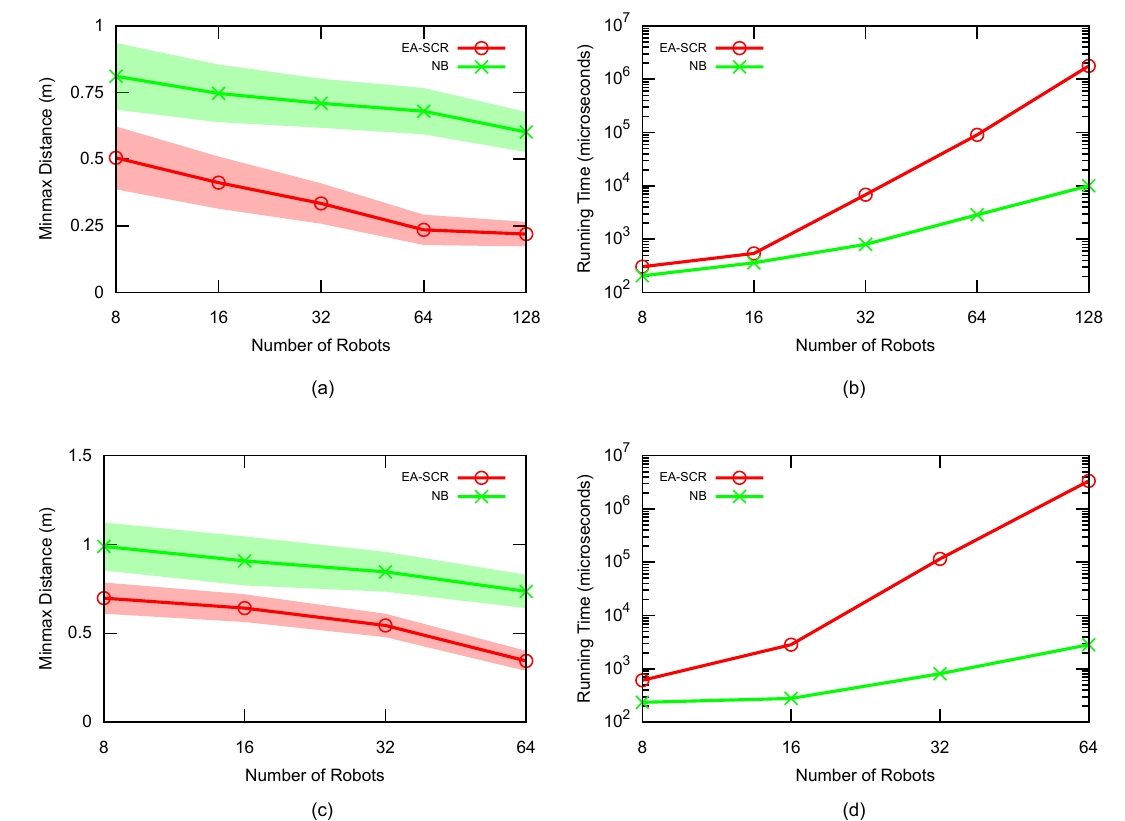}
\caption{Comparison of EA-SCR and NB algorithms. (a) $k=3$. metric: minmax distance. (b) $k=3$. metric: running time. (c) $k=4$. metric: minmax distance. (d) $k=4$. metric: running time.}
\label{figt}
\end{figure}

The running time of the EA-SCR algorithm is slightly higher than that of NB for lower values of $n$ and EA-SCR catches up with NB as $n$ grows. Both algorithms perform better than BT in terms of running time.

\textbf{Results for k=3 and k=4}: In this experiment, we compare EA-SCR and NB algorithms in terms of minmax distance and running time for $k=3$ and $k=4$. The experimental results are shown in Figure~\ref{figt}. Similar to the case of $k=2$, EA-SCR algorithm performs better than NB algorithm in terms of minmax distance. The running time of the EA-SCR algorithm is higher than NB, but still less than a few seconds for teams with 128 robots. The reason behind EA-SCR having a high running time is that testing for $k$-connectivity for high values of $k$ is computationally expensive as already explained in Section~\ref{eaalgo}. Although EA-SCR has a higher running time than NB, EA-SCR gives 30\% lower minmax distance compared to NB, thus providing a nice trade-off between running time and solution quality for higher values of $k$.

In this experiment, we do not compare with BT algorithm as it is only applicable for $k=2$. Also, for the sake of brevity, we do not present results for the total distance metric. EA-SCR performs better than NB in terms of the total distance metric similar to the results for $k=2$.

\subsection{Hardware Experiments}
\label{pmcf}
\begin{figure}[ht]
    \centering
    \subfloat[\label{fig:animation 1}]{
    \includegraphics[width=0.48\textwidth]{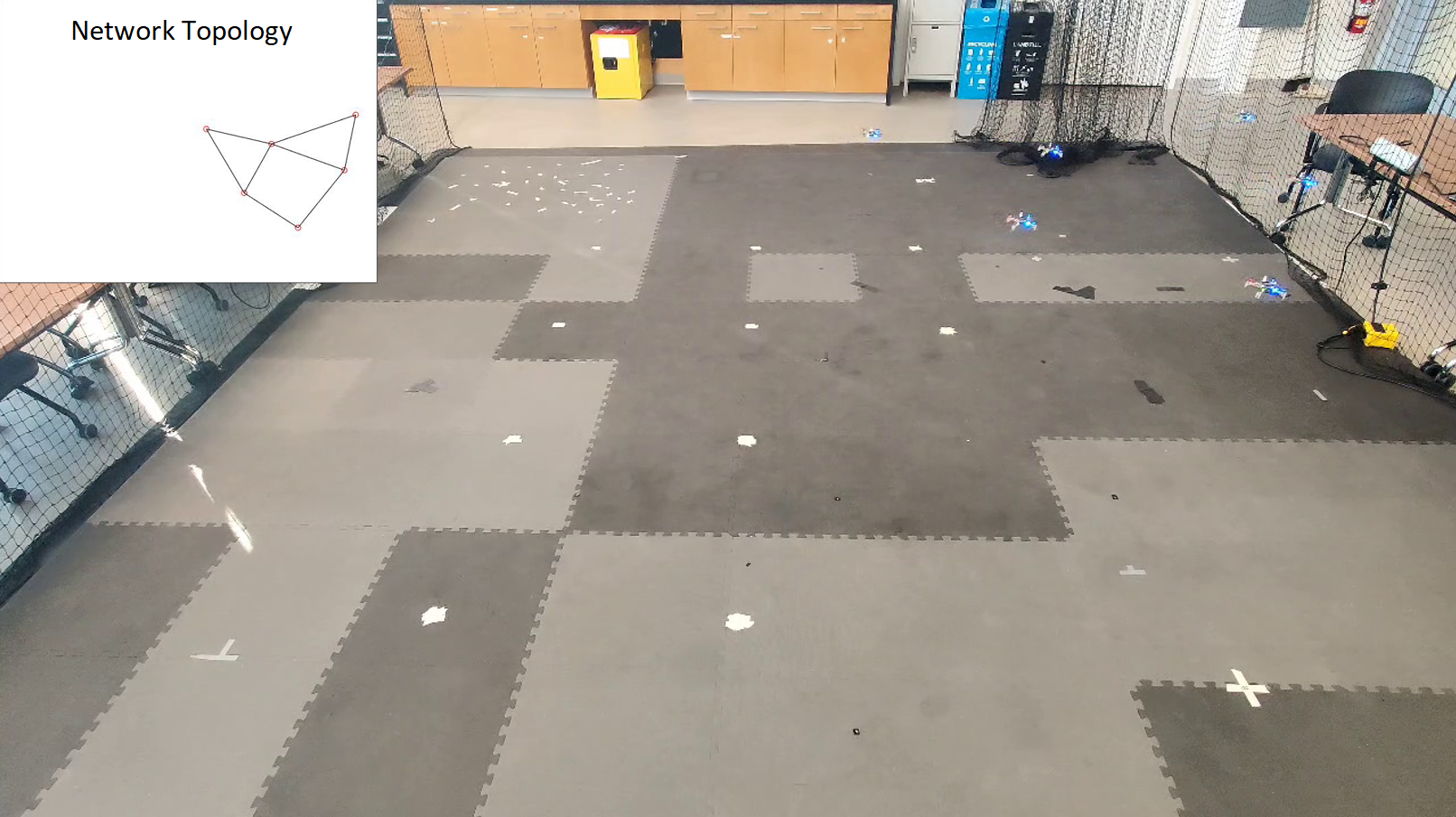}
    }
    \subfloat[\label{fig:animation 2}]{
    \includegraphics[width=0.48\textwidth]{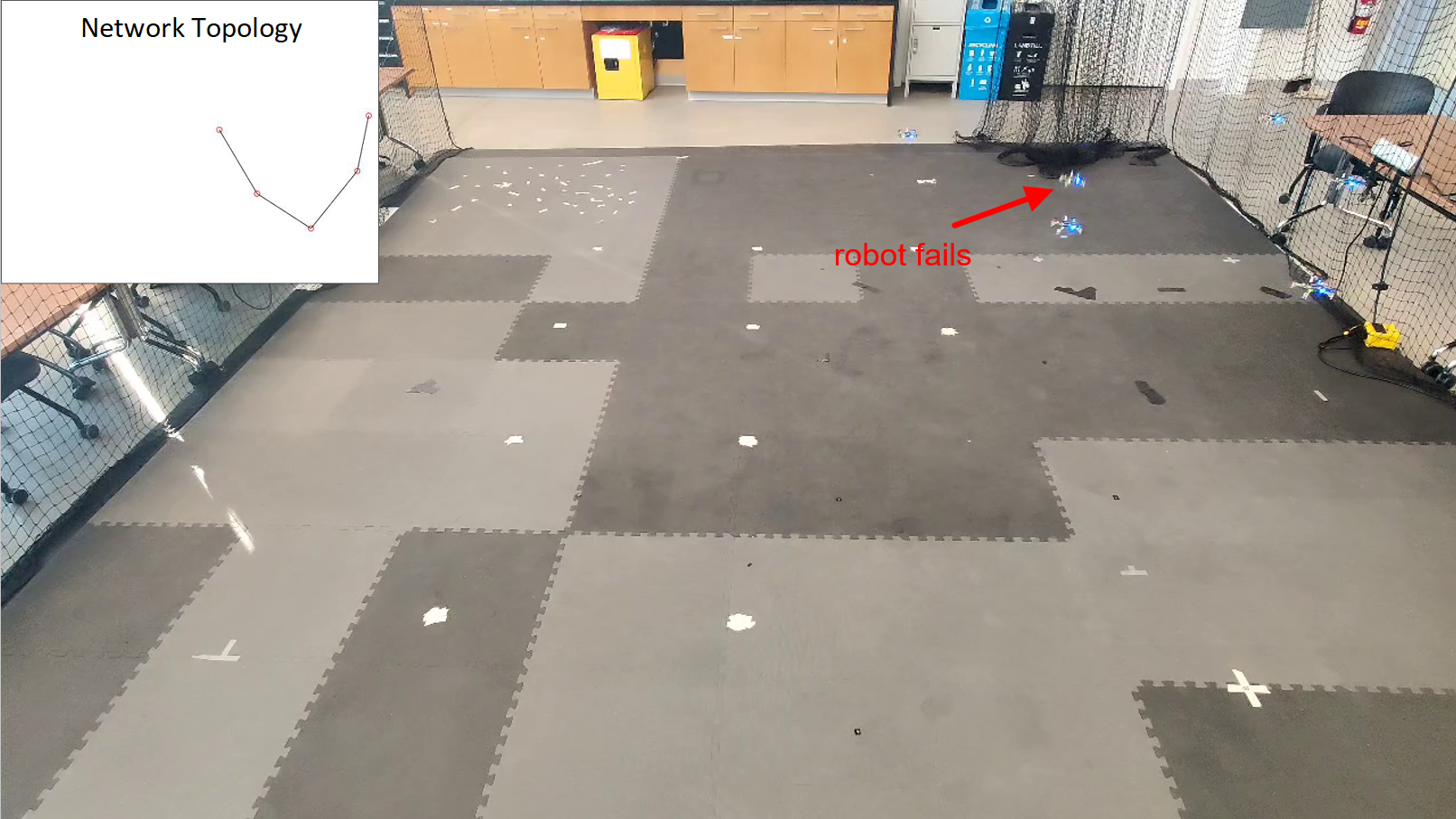}
    }\\
    \subfloat[\label{fig:animation 3}]{
    \includegraphics[width=0.48\textwidth]{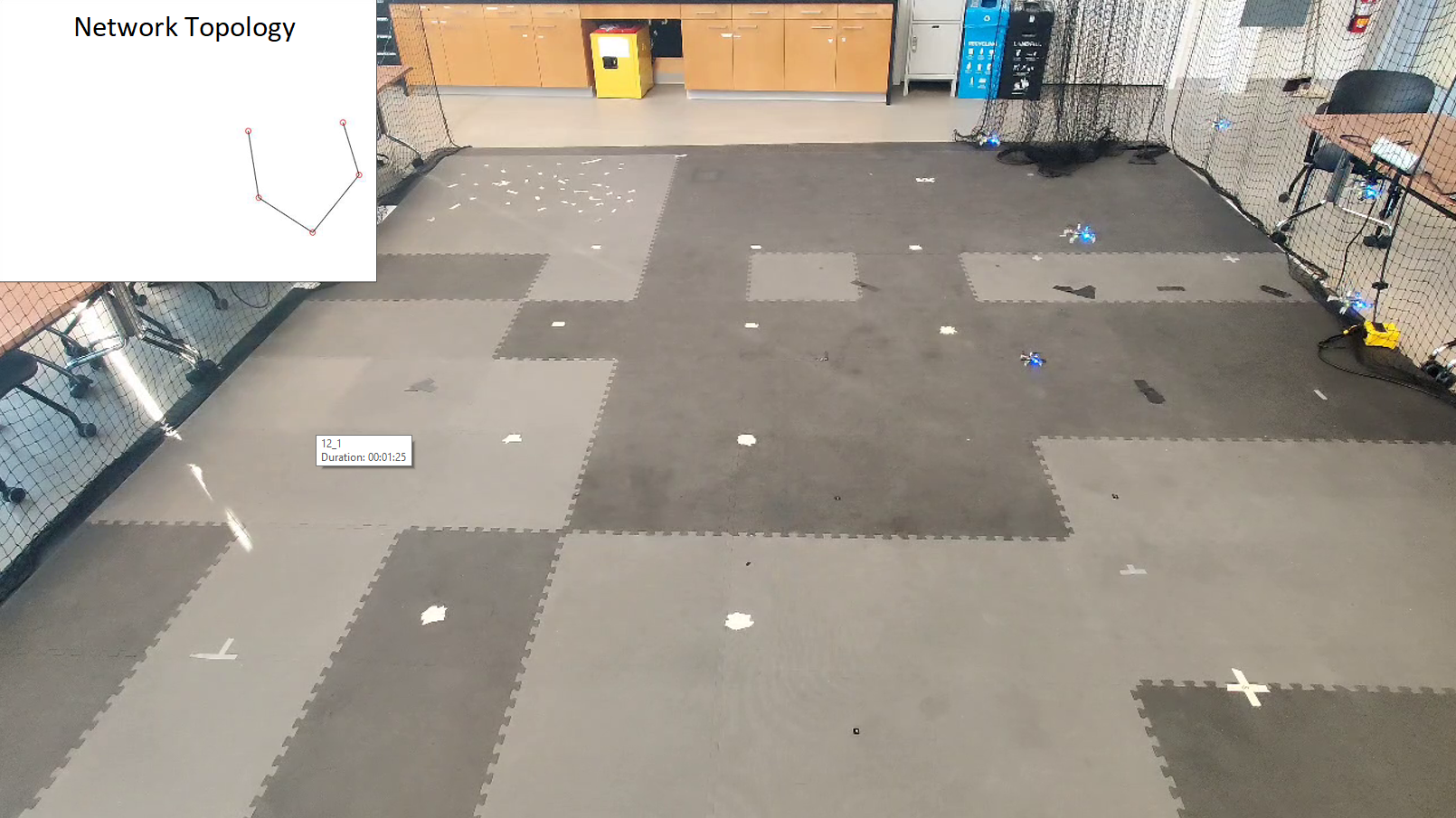}
    }
    \subfloat[\label{fig:animation 4}]{
    \includegraphics[width=0.48\textwidth]{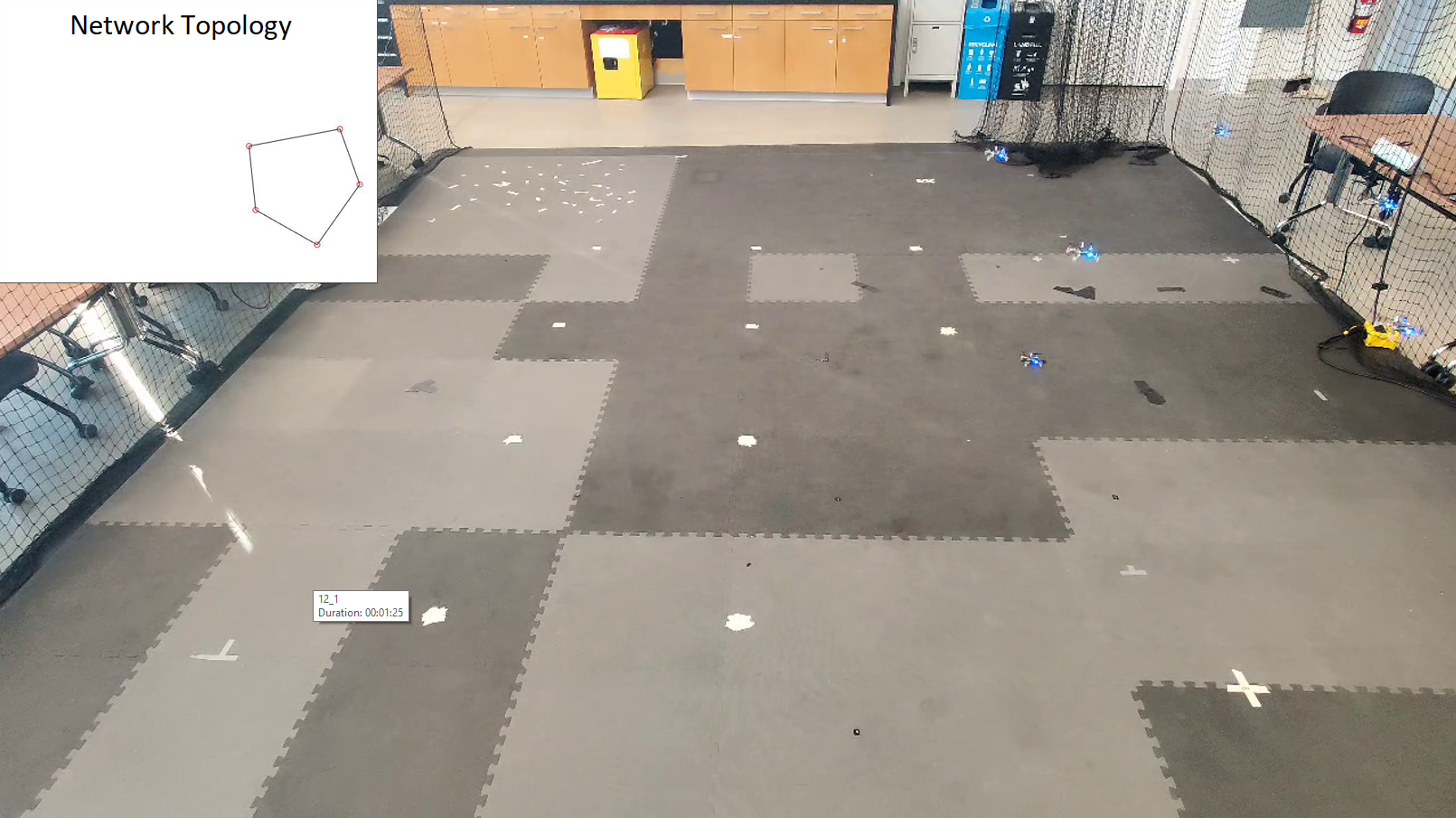}
    }
	\caption{Six CrazyFlies performing persistent monitoring task. The upper left window in each figure shows the network topology of the team.} 
	\label{fig:animation}
\end{figure}

We test the EA-SCR algorithm deployed in hardware for the Persistent Monitoring (PM) task~\cite{permon} using a setup similar to~\cite{rabban2021failure}. Persistent monitoring is a patrolling task which involves repeatedly visiting of the unoccupied space in the environment. We use six Crazyswarm nano-quadrotors \cite{preiss2017crazyswarm} to monitor a $4m \times 4m$ space with $h=1m$ and $k=2$. As shown in Fig. \ref{fig:animation}, the CrazyFlie robots maintain a 2-connected graph while executing the PM task (Fig. \ref{fig:animation 1}). When a robot fails resulting in loss of 2-connectivity (Fig. \ref{fig:animation 2}), the robots move towards new positions (calculated using EA-SCR algorithm) to regain 2-connectivity (Fig. \ref{fig:animation 3}). 2-connectivity is restored when the robots reach their goals (Fig. \ref{fig:animation 4}). Please refer to Rabban et al.~\cite{rabban2021failure} for more details about the experimental setup. A video of the system in operation can be found here: \texttt{http://youtu.be/AYcu3Itwqc4}.

%% file: 07_conclusion.tex
\section{Conclusion}
\label{con}

In this work, we have proposed a scalable algorithm for the FCR problem which has outperformed the existing algorithms. We have also developed a QCP formulation to solve the FCR problem optimally. We have demonstrated the effectiveness of our proposed solution by conducting empirical studies. 

In the future, we intend to prove the approximability of the performance of EA-SCR algorithm. We also plan to work on the FCR variant in which the environment contains obstacles. One possible way to deal with obstacles is to discretize the environment and solve a discrete version of FCR. Another direction is to devise the distributed counterpart of the proposed algorithm.